\documentclass[final]{cvpr}

\usepackage{times}
\usepackage{epsfig}
\usepackage{graphicx}
\usepackage{amsmath}
\usepackage{amssymb}

\usepackage{mathrsfs}
\usepackage{url}
\usepackage{enumitem}
\usepackage{subfigure}
\usepackage{algpseudocode}
\usepackage{setspace}
\usepackage{booktabs}
\usepackage{enumitem}
\usepackage{multirow}
\usepackage[linesnumbered,algoruled,boxed,lined]{algorithm2e}

\newcommand{\figdir}{figures}

\usepackage[pagebackref=false,breaklinks=true,letterpaper=true,colorlinks,urlcolor=blue,citecolor=blue,linkcolor=blue,bookmarks=false]{hyperref}

\def\cvprPaperID{4215} 
\def\confYear{CVPR 2021}

\begin{document}

\title{IoU Attack: Towards Temporally Coherent Black-Box \\ Adversarial Attack for Visual Object Tracking}

\author{
Shuai Jia$^1$\quad 
Yibing Song$^2$\quad 
Chao Ma$^1$\thanks{Corresponding author.}\quad
Xiaokang Yang$^1$ \\
$^1$MoE Key Lab of Artificial Intelligence, AI Institute, Shanghai Jiao Tong University  \\
$^2$Tencent AI Lab\\
{\tt\small {\{jiashuai,chaoma,xkyang\}@sjtu.edu.cn, yibingsong.cv@gmail.com}}
}

\maketitle

\thispagestyle{empty}
\pagestyle{empty}


\begin{abstract}

Adversarial attack arises due to the vulnerability of deep neural networks to perceive input samples injected with imperceptible perturbations. Recently, adversarial attack has been applied to visual object tracking to evaluate the robustness of deep trackers. Assuming that the model structures of deep trackers are known, a variety of white-box attack approaches to visual tracking have demonstrated promising results. However, the model knowledge about deep trackers is usually unavailable in real applications. In this paper, we propose a decision-based black-box attack method for visual object tracking. In contrast to existing black-box adversarial attack methods that deal with static images for image classification, we propose IoU attack that sequentially generates perturbations based on the predicted IoU scores from both current and historical frames. By decreasing the IoU scores, the proposed attack method degrades the accuracy of temporal coherent bounding boxes (i.e., object motions) accordingly. In addition, we transfer the learned perturbations to the next few frames to initialize temporal motion attack. We validate the proposed IoU attack on state-of-the-art deep trackers (i.e., detection based, correlation filter based, and long-term trackers). Extensive experiments on the benchmark datasets indicate the effectiveness of the proposed IoU attack method. The source code is available at {\small{\url{https://github.com/VISION-SJTU/IoUattack}}}.

\end{abstract}

\vspace{-0.15in}

\renewcommand{\tabcolsep}{1pt}
\def\swone{0.8\linewidth}
\def\swtwo{0.48\linewidth}
\def\swfour{0.5\linewidth}

\begin{figure}[t]
    \small
	\begin{center}
		\includegraphics[width=\swone]{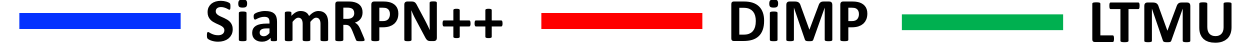}\\
		\vspace{2mm}
			\begin{tabular}{cc}
				\includegraphics[width=\swtwo]{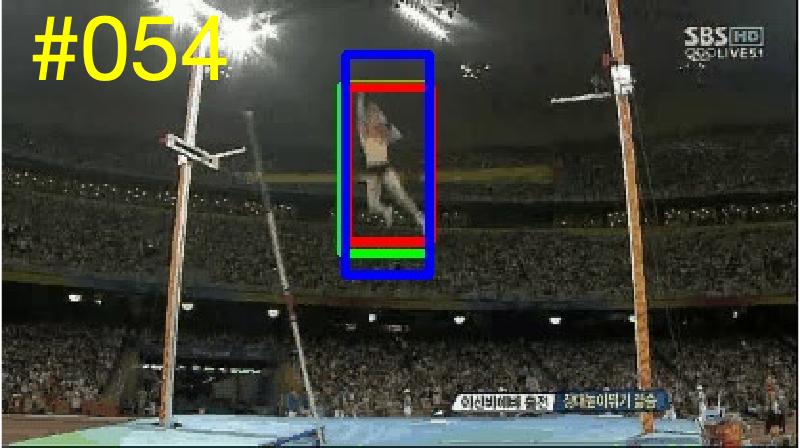}&
				\includegraphics[width=\swtwo]{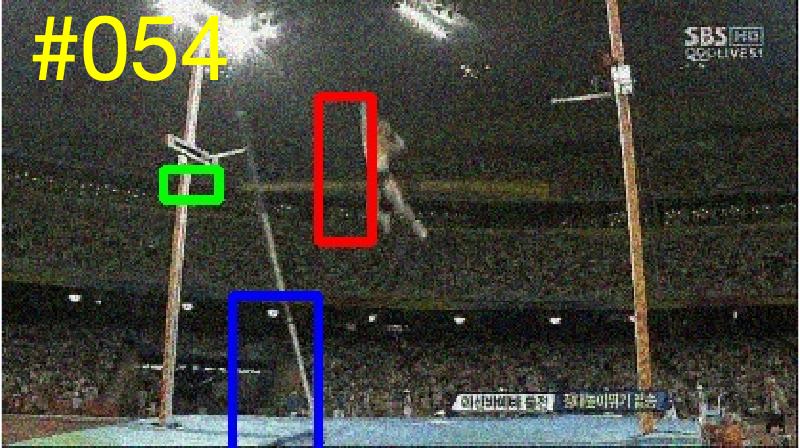}\\
				\includegraphics[width=\swtwo]{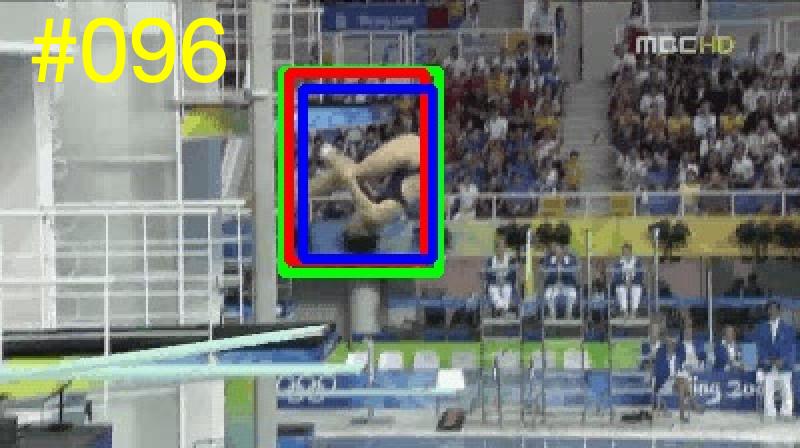}&
				\includegraphics[width=\swtwo]{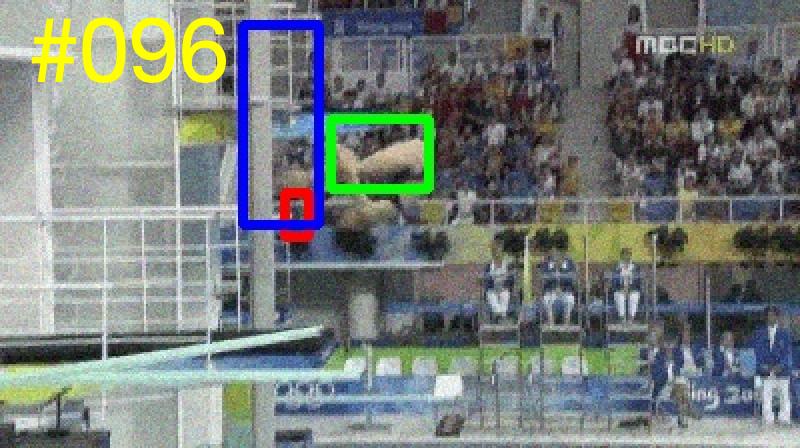}\\
				\includegraphics[width=\swtwo]{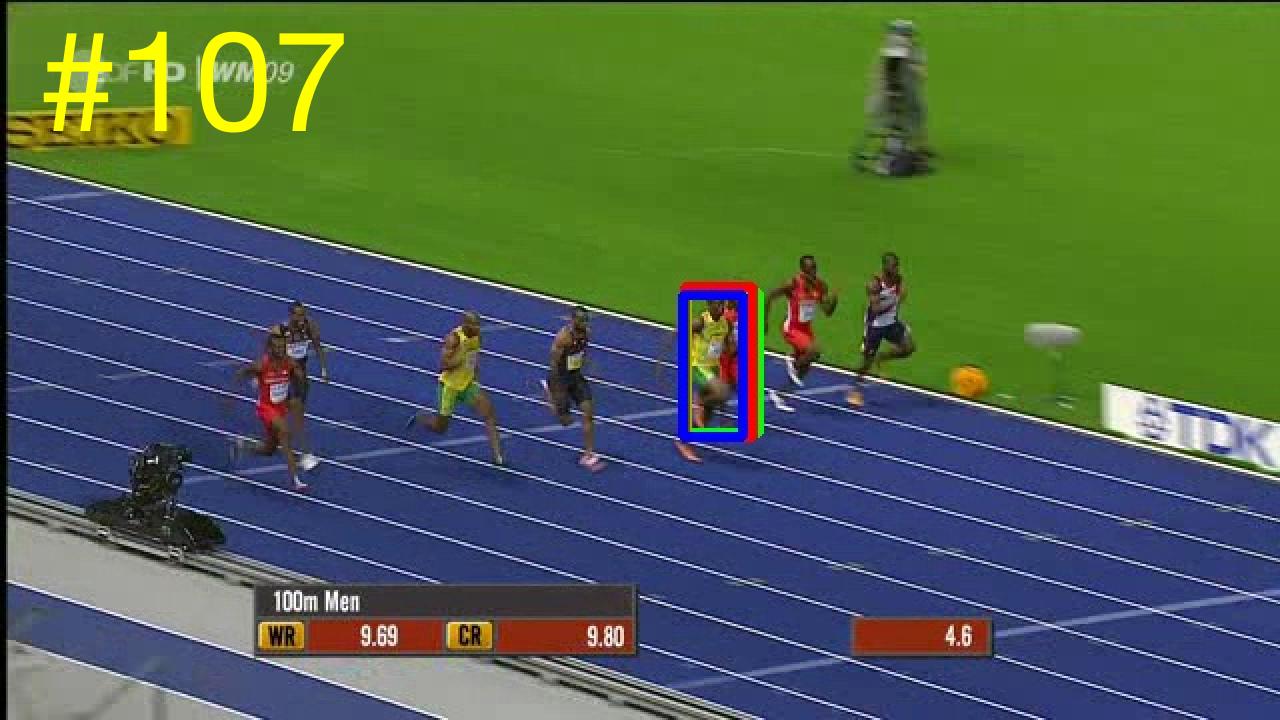}&
				\includegraphics[width=\swtwo]{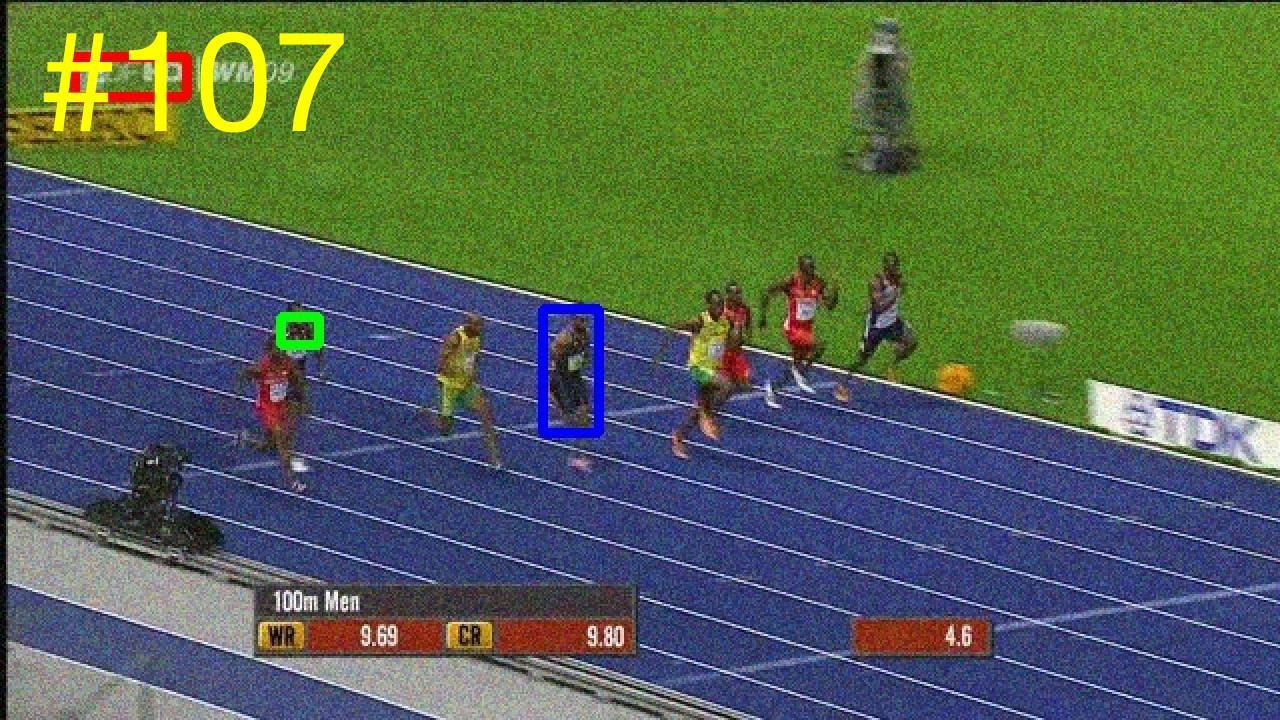}\\
				(a) Original results &(b)  IoU attack\\
			\end{tabular}
		\vspace{-0.1in}
	\end{center}
\caption{IoU attack for visual object tracking. State-of-the-art deep trackers (i.e., SiamRPN++~\cite{li-cvpr19-siamrpn++}, DiMP~\cite{bhat-iccv19-learning}, and LTMU~\cite{dai-cvpr20-high}) effectively locate target objects in the original video sequences as shown in (a). Our IoU attack decreases their tracking accuracies by injecting imperceptible perturbations as shown in (b).}
\label{fig:intro}
\vspace{-0.1in}
\end{figure}

\section{Introduction}

\renewcommand{\tabcolsep}{0pt}

Visual object tracking is one of the fundamental computer vision problems with a wide range of applications. The convolutional neural networks (CNNs) have significantly advanced visual tracking performance. Meanwhile, the enigma of interpreting CNNs has perplexed existing visual tracking algorithms as well. For example, injecting imperceptible perturbations into input images leads deep neural networks to predict incorrectly~\cite{Szegedy-iclr14-aa,xie-cvpr19-aad,zhao-cvpr20-attack}. To investigate the robustness of visual tracking algorithms with deep models, recent approaches~\cite{chen-cvpr20-one,yan-cvpr20-cooling,jia-eccv20-robust,liang-eccv20-efficient} assume that the model structures of deep tracking algorithms are known and carry out white-box attack on them. Despite the demonstrated promising results, the concrete structures and parameters of deep trackers are barely known in real applications. In this paper, we investigate black-box adversarial attack for visual tracking, where the model knowledge of deep trackers is unknown. 

Prevalent black-box attack algorithms inject imperceptible perturbations into input images to decrease network classification accuracies. Although these methods are effective to attack static images, they are not suitable to attack temporally moving objects in videos. This is because deep trackers maintain temporal motions of the target object within tracking models (i.e., the correlation filters~\cite{danelljan-cvpr17-eco,song-iccv17-crest} or deep binary classifiers~\cite{nam-cvpr16-mdnet,jung-eccv18-rtmdnet,li-cvpr18-siamrpn,li-cvpr19-siamrpn++}). When localizing the target object, these deep trackers produce temporally coherent bounding boxes (bbxs). Meanwhile, deep trackers constrain the search area to be close to the predicted bbx from the last frame. As existing black-box methods rarely degrade temporally coherent bbxs, perturbations produced based on CNN classification scores are not effective for visual tracking. An intriguing direction thus arises to investigate the black-box attack on both individual frames and temporal motions among sequential frames with a holistic decision-based approach.

In this paper, we propose IoU attack for visual tracking. IoU attack is a decision-based black-box attack method which focuses on both image content and target motions in video sequences. When processing each frame, we start image content attack with two bbxs. One is predicted by the deep tracker using the original frame, which is perturbation free. The other one is predicted by the same tracker using the same frame with noisy perturbations. These two bbxs are used to compute an IoU score as feedback to our IoU attack. For each frame, we use an iterative orthogonal composition method for image content attack. During each iteration of orthogonal attack, we first randomly generate several tangential perturbations whose noise levels are the same. Then, we compute their IoU scores and select the tangential perturbation with the lowest score. The selected perturbation is the most effective one to attack the current frame at the current iteration. We then increase the selected perturbation in its normal direction to add a small amount of noise, which is the normal perturbation. We compose both tangential and normal perturbations to generate the perturbations for the current iteration of orthogonal attack. 

For target motion attack, we compute an IoU score between the bbxs from both the current and the previous frames. This IoU score is integrated into the tangential perturbation identification process. To this end, our orthogonal attack deviates a deep tracker from its original performance of both the current and historical frames. We transfer the learned perturbations to the next few frames as perturbation initialization to reinforce temporal motion attack. As a result, the deviation from the original tracking results ensures the success of black-box attack on deep trackers shown in Figure~\ref{fig:intro}. We extensively validate the proposed IoU attack on state-of-the-art methods including detection based~\cite{li-cvpr19-siamrpn++}, correlation filter based~\cite{bhat-iccv19-learning}, and long-term~\cite{dai-cvpr20-high} trackers. Experiments on benchmark datasets demonstrate the effectiveness of the proposed black-box IoU attack.

\section{Related Work}

\renewcommand{\tabcolsep}{0pt}

In this section, we briefly introduce recent state-of-the-art trackers and their basic principles. Besides, we also review recent adversarial attack methods, especially for the aspect of black-box attack.

\subsection{Visual Object Tracking}

Visual object tracking has received widespread attention in the last decade and brings about a series of new benchmark datasets~\cite{OTB-2015,nfs,uav123,trackingnet,lasot}. Existing trackers can be generally categorized as offline trackers and online update trackers. Offline trackers do not update their model parameters during the inference, leading to a higher speed. These trackers consider tracking as a discriminative object detection problem. They generate candidate regions and classify the target or background to locate. Bounding box regression~\cite{fasterrcnn} is always used to locate precisely. Among them, siamese based methods ~\cite{li-cvpr18-siamrpn,li-cvpr19-siamrpn++,wang-cvpr19-udt,zhang-cvpr19-siamdw,guo-cvpr20-siamcar,chen-cvpr20-siamese,voigtlaender-cvpr20-siam,wang-ijcv21-udt} are typical structures consisting of a template branch and a search branch. SiamRPN~\cite{li-cvpr18-siamrpn} draws a region proposal network to formulate a one-shot detection by comparing the similarity between two branches. SiamRPN++~\cite{li-cvpr19-siamrpn++} applies a deeper network ResNet instead of commonly-used AlexNet to improve the tracking accuracy and maintain the real-time speed. 

Online update trackers constantly update their models during the inference to adapt to the current scenarios~\cite{pan2021videomoco}. MDNet~\cite{nam-cvpr16-mdnet,song-cvpr18-vital,pu-nips18-dat} regard tracking as a classification to distinguish the target and background. During the inference, they collect the samples from previous frames to enhance the target appearance. UpdateNet~\cite{zhang-iccv19-learning} formulates an update strategy into siamese based trackers to maintain the temporal motion between frames. Besides, correlation filter based methods also belong to online update trackers. They typically learn the discriminative correlation filter by deep or hand-craft features to estimate the target location. Recently, DiMP~\cite{bhat-iccv19-learning} learns a discriminative learning loss to exploit both target and background appearance information for target model prediction. PrDiMP~\cite{danelljan-cvpr20-probabilistic} proposes a probabilistic regression formulation to address the modeling label noise. 

Furthermore, existing long-term trackers~\cite{yan-iccv19-skimming,dai-cvpr20-high} integrate an online update module to improve the tracking performance. The re-detection module is mostly introduced to handle the disappearance and reappearance of the target, involving more challenges into adversarial attack. LTMU~\cite{dai-cvpr20-high} is a long-term tracker with a meta-updater, which learns to guide the tracker's update to gain helpful appearance information for accuracy. In this work, we implement our adversarial attack on three representative trackers~\cite{li-cvpr19-siamrpn++,bhat-iccv19-learning,dai-cvpr20-high} to illustrate the generality of our black-box attack method. 

\subsection{Adversarial Attack}

Convolution Neural Networks (CNNs) have been deployed in various tasks of computer vision today. However, recent studies~\cite{goodfellow-iclr15-explaining,Szegedy-iclr14-aa} notice that CNNs are sensitive to the imperceptible perturbations in adversarial examples. The intentional light-weight perturbations deteriorate the performance dramatically. Existing adversarial attack methods~\cite{goodfellow-iclr15-explaining,moosavi-cvpr16-deepfool,ilyas2-icml18-black,dong-cvpr19-aa,qi-iclr21-smia} mainly focus on static image tasks like classification, segmentation and detection.  Except for attacking digital images, some studies implement physical attacks~\cite{eykholt-cvpr18-robust,wiyatno-iccv19-physical} in concrete applications (e.g., autonomous driving). They generate a distractor in the real world to cause CNNs models to misclassify or fail to detect, leading to a security problem.

Overall, existing adversarial attack methods are mainly divided into two categories: white-box and black-box attack. In white-box attack, the adversary assumes to gain all knowledge of the attacked target, such as the learned parameters, the concrete structure, etc.  Compared with white-box attack, black-box attack has limited knowledge of the model but is closer to the practical scenarios. It is often modeled on querying the method by inputs, acquiring the final labels or confidence scores. Black-box attack roughly fails into transfer-based, score-based, and decision-based attack~\cite{dong-cvpr19-efficient}. Transfer-based attack~\cite{liu-Iclr17-delving} utilizes the transferability of adversarial examples generated by white-box models. Score-based attack knows the predicted probability of classification, relying on approximated gradients to generate adversarial examples~\cite{ilyas2-icml18-black}. In decision-based attack, only the final label of classification is accessible~\cite{brendel2017decision} to the threat model. For black-box attack in visual object tracking, we assume that only the outputs of trackers (i.e., predicted bounding boxes) are available.

In the field of visual object tracking, some methods ~\cite{chen-cvpr20-one,yan-cvpr20-cooling,jia-eccv20-robust,liang-eccv20-efficient,guo-eccv20-spark} explore the adversarial attack on different trackers.  Chen et al.~\cite{chen-cvpr20-one} propose a one-shot adversarial attack method by optimizing the batch confidence loss and the feature loss to modify the initial target patch. Yan et al.~\cite{yan-cvpr20-cooling} design an adversarial loss to cool the hot regions on the heatmaps and shrink the predicted bounding box. All adversarial attack methods motioned above are summarized as white-box attack. In the real world, it is hard to know the concrete knowledge of trackers, causing the white-box attack methods less practical. In this work, we propose a novel decision-based black-box adversarial attack method for visual tracking. Motion consistency is taken into our attack method to further deteriorate the tracking performance.

\section{Proposed Method}

\renewcommand{\tabcolsep}{0pt}

\begin{figure}[t]
    \small
	\begin{center}
		\begin{tabular}{c}
			\includegraphics[width=1\linewidth]{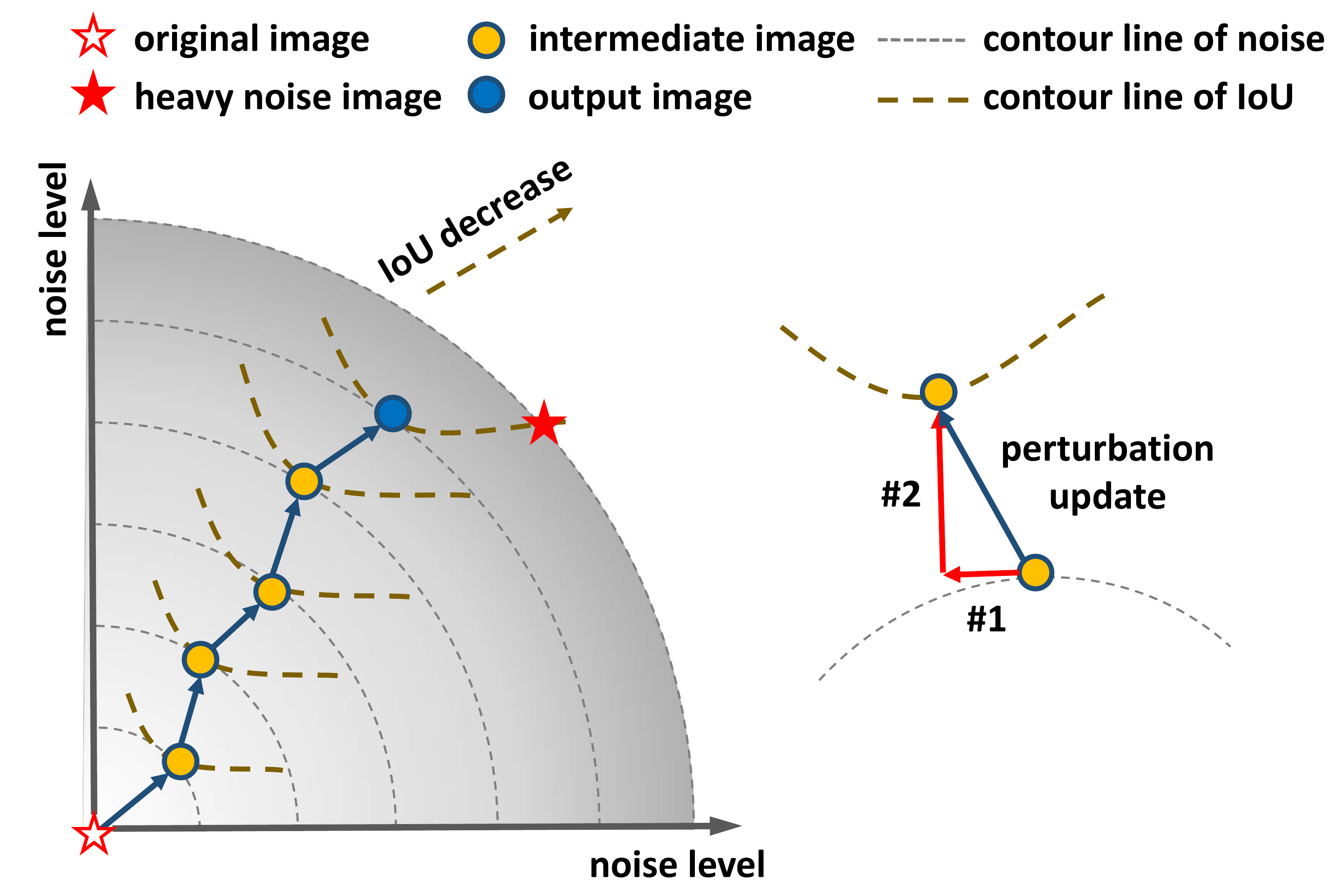}\\
		\end{tabular}
		\begin{tabular}{cc}
			\ \ \ (a) Iterative perturbation update \  & \ \ \ \ (b) Orthogonal composition\\
		\end{tabular}
		\vspace{-0.1in}
	\end{center}
	\caption{An intuitive view of IoU attack in the image space. In (a), we show that the increase of noise level positively correlates to the decrease of IoU scores but their directions are not exactly the same. The IoU attack method iteratively finds the intersection points (i.e., intermediate images) between each contour line of noise increase and IoU decrease. These intermediate images gradually decrease IoU scores with the lowest amount of noise. In (b), we show the orthogonal composition during each iteration. We generate noise hypothesis tangentially according to the current contour line (i.e., \#$1$) and increase a small amount of noise in the normal direction (i.e., \#$2$). The intersection point will be identified from the hypothesis that yields the lowest IoU at the same noise level. The updated perturbation in each iteration is the composition of \#$1$ and \#$2$.}\label{fig:algo}
\end{figure}

The proposed IoU attack aims to gradually decrease IoU scores (i.e., bbx overlap values) during iterations by using the minimal amount of noise. Figure~\ref{fig:algo} shows an intuitive view of the proposed IoU attack.

\def\swtwo{0.45\linewidth}

\subsection{Motivation}

Current studies on black-box attack mainly focus on static image recognition while the temporal motions of visual tracking are untouched. This is limited to attack deep trackers as the target object motion is maintained temporally. Meanwhile, deep trackers utilize temporally coherent schemes (i.e., search region constraint, and online update) to ensure tracking accuracy. The image content and temporal motions are equally important for black-box attack on visual tracking.

The proposed IoU attack is to make the prediction results of one tracker deviate from its original performance. This is because of the tracking scenario where there is only one ground-truth bounding box (bbx) available (i.e., bbx annotation on the first frame). We define the original performance of one tracker is that it predicts one bbx on each frame without noise addition. By adding heavy-noisy perturbations, we make the same tracker predict another bbx and compute the spatial IoU score based on these two bbxs. Meanwhile, we use the bbx from the current frame and the one from the previous frame to compute a temporally coherent IoU score, which is then fused with the spatial IoU score. As state-of-the-art trackers demonstrate premier performance on the benchmarks, gradually decreasing the IoU scores by involving consecutive video frames indicates that their tracking performance deteriorates significantly. The IoU measurement suits different trackers as long as they predict one bbx for each frame.

\subsection{IoU Attack}

Figure~\ref{fig:algo} shows an intuitive view of how the proposed IoU attack gradually decreases the IoU scores between frames. Given a clean input frame, we first add heavy uniform noise on it to generate a heavy noise image where the IoU score is low. Along the direction from the clean image to the heavy noise image, the IoU scores gradually decrease while the noise level increases. The direction of IoU decrease positively correlates to that of noise increase but they are not exactly the same. IoU attack aims to progressively find a decreased IoU score while introducing the lowest amount of noise. The contour lines of IoU shown in Figure~\ref{fig:algo}(a) indicate the tracker performance with regard to different noise perturbations, which can not be explicitly modeled in practice. From another perspective, IoU attack aims to identify one specific noise perturbation leading to the lowest IoU score among the same amount of noise levels. The identification process is fulfilled by orthogonal composition illustrated as follows.

\begin{figure*}[t]
	\centering
	\includegraphics[width=1\linewidth]{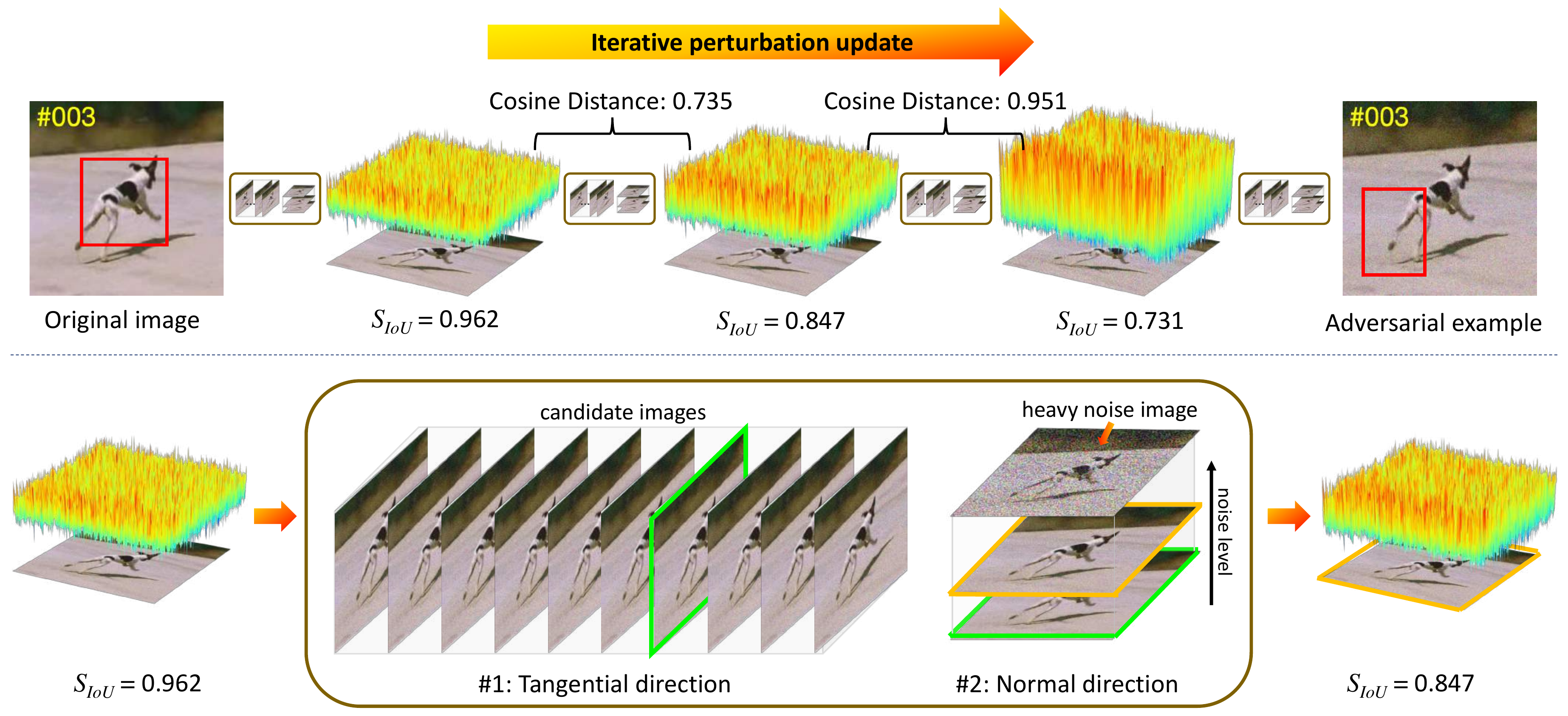}
	\vspace{-3mm}
	\caption{Variations of adversarial perturbations during IoU attack. The 3D response map above the image represents the difference between the original image and the adversarial example at different IoU scores. The IoU score decreases as the magnitude of perturbations increases. The variations of perturbations are illustrated in the first row. The orthogonal composition is shown in the second row, including tangential direction and normal direction. The image framed in green represents the minimal IoU score in the tangential direction and the image framed in yellow represents the moving step towards the heavy noise image.
	} \label{fig:method}
\end{figure*}

We denote the original image on the $t$-th frame as $I_0$, the heavy noise image as $H$, and the intermediate image on the $k$-th iteration as $I_k$. In the ($k$+1)-th iteration, we first randomly generate several Gaussian distribution noise $\eta\sim\mathcal N(0,1)$ and select the tangential perturbation $\eta$ from $n$ of them as:
\begin{equation}\label{eq:tan}
d(I_0,I_k)=d(I_0,I_k+\eta),
\end{equation}
where $d$ is the pixel-wise distance measurement between two images. Eq.~\ref{eq:tan} ensures tangential perturbations at the same noise level. The selected $\eta^j$ ($j\in[1,2,...,n]$) is the perturbation tangential towards the contour line of noise level at the point $I_k$. We generate one $I^j$ ($j\in[1,2,...,n]$) according to each $\eta^j$ and use the tracker to predict a bbx $B_t^j$ on it. Then, we define the IoU score $S_{\rm IoU}$ as:
\begin{equation}\label{eq:score}
S_{\rm IoU} = \lambda \cdot S_{\rm spatial} + (1-\lambda) \cdot S_{\rm temporal},
\vspace{-0.05in}
\end{equation}
\begin{align}
S_{\rm spatial \ \ \ } &= \frac{B^{\rm orig}_t\cap B_t^j}{B^{\rm orig}_t\cup B_t^j}, \\
S_{\rm temporal} &= \frac{B^{\rm orig}_{t-1}\cap B_t^j}{B^{\rm orig}_{t-1}\cup B_t^j}, 
\end{align}
where $S_{\rm spatial}$ denotes the spatial IoU score between the predicted bbx $B_t^j$ and the original noise-free bbx $B_t^{\rm orig}$ at the $t$-th frame, $S_{\rm temporal}$ denotes the temporal IoU score with the original noise-free bbx $B_{t-1}^{\rm orig}$ at the ($t$-1)-th original frame, and $\lambda$ is the scalar to balance the influence of spatial and temporal IoU scores. We attack $S_{\rm IoU}$ to perform both image content attack and temporal motion attack. In total, we obtain $n$ IoU scores and select $I^j$ whose $S_{\rm IoU}$ is lowest. An example of $\eta^j$ is visualized as \#$1$ in Figure~\ref{fig:algo}(b).

After getting the tangential perturbation, we denote $\eta^j$ as neighboring hypothesis based on $I_k$ and make $I_k+\eta^j$ towards the heavy noise image $H$ as:
\begin{equation}\label{eq:normal}
I_{k+1}^j=(I_k+\eta^j)+\epsilon\cdot \psi(H,I_k+\eta^j),
\end{equation}
where $\epsilon$ controls the moving step towards $H$ and $\epsilon\cdot\psi(H,I_k+\eta^j)$ is the perturbation following the noise increase direction (i.e., normal direction towards the contour line of noise level). We adjust the parameter $\epsilon$ moderately to limit the variation of perturbations.  An example of $\epsilon\cdot\psi(H,I_k+\eta^j)$ is visualized as \#$2$ in Figure~\ref{fig:algo}(b). To this end, $I_{k+1}^j$ is the intermediate image on the ($k$+1)-th iteration, consisting of the composed perturbations from both tangential and normal directions. We continuously perform the iteration until the IoU score is below the predefined threshold or the perturbations exceed the maximum. We transfer the learned perturbations $P_t$ to the next few frames. The learned perturbations become the initialized perturbations, which are added on $I_{0}$ of ($t$+1)-th frame to encode temporal motion attack from previous frames. The pseudo code of the black-box IoU attack is shown in Algorithm~\ref{algo:1}.

\begin{algorithm}[t]
    \label{algo:1}
	\caption{Black-box IoU Attack}
	\KwIn{
		\hspace*{0.08in}Input video $V$ with $M$ frames;\\ 
		\hspace*{0.53in}Initialization perturbations $P_1=0$;\\
		\hspace*{0.53in}Target bbx $B_1$ on the first frame;}
	\KwOut{Adversarial examples of $M$ frames; }
	\For{$t = 2 $  \KwTo $M$}
	{
		Get current frame $I_0$ and predict bbx $B^{\rm orig}_t$ \;
		\textit{$I_0=I_0+ \alpha \cdot P_{t-1}$}\;
		\For{$k = 0$ \KwTo $K$-$1$}
		{
			\tcp{Tangential direction}
			Generate $N$ random perturbations $\eta$\; 
			Select $n$ of them according to Eq.~\ref{eq:tan}\;
			\For{$j = 1$ \KwTo $n$}
			{   
			    Predict the bbx $B^j_t$ on $I_k+\eta^j$ \;
			    Compute $S_{\rm IoU}$ according to Eq.~\ref{eq:score}\;
			}
			Identify $j$ whose $S_{\rm IoU}$ is lowest\;
			\tcp{Normal direction}
			Adjust $\epsilon$ to decrease $S_{\rm IoU}$\;
		    Generate $I^j_{k+1}$ according to Eq.~\ref{eq:normal}\;
		}
		Obtain learned perturbations $P_t=I^j_{k+1}-I_0$\;
		\Return $I^j_{K}$;
	}
\end{algorithm}

\begin{table*}[ht]
	\small
	\begin{center}
		\caption{Comparison of tracking results with original sequences, random noise, and IoU attack of SiamRPN++~\cite{li-cvpr19-siamrpn++}, DiMP~\cite{bhat-iccv19-learning} and LTMU~\cite{dai-cvpr20-high}  respectively on the VOT2019~\cite{vot2019} dataset.} \label{table:table1}
		\vspace{0.05in}
		\begin{tabular*} {17cm} {@{\extracolsep{\fill}}lcccccccccccc}
			\toprule
			\multirow{2}*{\ \ Trackers} & \multicolumn{3}{c}{ Accuracy $\uparrow$ }  & \multicolumn{3}{c}{ Robustness $\downarrow$ } & \multicolumn{3}{c}{ Failures $\downarrow$ } & \multicolumn{3}{c}{ EAO $\uparrow$ } \\
			\cmidrule(r){2-4} \cmidrule(r){5-7} \cmidrule(r){8-10} \cmidrule(r){11-13} 
			~ &Orig. &Rand. &Attack &Orig. &Rand. &Attack &Orig. &Rand. &Attack &Orig. &Rand. &Attack\\
			\midrule
			\ \ SiamRPN++  &0.596 & 0.591 & \textbf{0.575} & 0.472 & 0.727 &\textbf{1.575} & 94 \ \ & 145 & \textbf{314} & 0.287 & 0.220 & \textbf{0.124}\\
			\ \ DiMP &0.568 & 0.567 &\textbf{0.474} & 0.277 & 0.373 & \textbf{0.641} & 55 \ \ & 74 \ \  & \textbf{127} & 0.332 & 0.284 & \textbf{0.195}\\
			\ \ LTMU &0.625 & 0.623 & \textbf{0.576} & 0.913 & 1.073 & \textbf{1.470} & 182 &214 & \textbf{293} &0.201 & 0.175 & \textbf{0.150}\\
			\bottomrule
		\end{tabular*}
	\end{center}
	\vspace{-0.2in}
\end{table*}

\begin{table*}[ht]
	\small
	\begin{center}
		\caption{Comparison of tracking results with original sequences, random noise, and IoU attack of SiamRPN++~\cite{li-cvpr19-siamrpn++}, DiMP~\cite{bhat-iccv19-learning} and LTMU~\cite{dai-cvpr20-high} respectively on the VOT2018~\cite{vot2018} dataset.} \label{table:table2}
		\vspace{0.05in}
		\begin{tabular*} {17cm} {@{\extracolsep{\fill}}lcccccccccccc}
			\toprule
			\multirow{2}*{\ \ Trackers} & \multicolumn{3}{c}{ Accuracy $\uparrow$ }  & \multicolumn{3}{c}{ Robustness $\downarrow$ } & \multicolumn{3}{c}{ Failures $\downarrow$ } & \multicolumn{3}{c}{ EAO $\uparrow$ } \\
			\cmidrule(r){2-4} \cmidrule(r){5-7} \cmidrule(r){8-10} \cmidrule(r){11-13} 
			~ &Orig. &Rand. &Attack &Orig. &Rand. &Attack &Orig. &Rand. &Attack &Orig. &Rand. &Attack\\
			\midrule
			\ \ SiamRPN++  & 0.602 & 0.587 & \textbf{0.568} & 0.239 & 0.365 & \textbf{1.171} & 51 \ \  & 78 \ \  & \textbf{250} &0.413 &0.301 &\textbf{0.129}\\
			\ \ DiMP & 0.574 & 0.560 & \textbf{0.507} &0.145 & 0.202 & \textbf{0.400} & 31 \ \ & 43 \ \  & \textbf{85} \ \  & 0.427 & 0.363 & \textbf{0.248}\\
			\ \ LTMU &0.624 & 0.622 & \textbf{0.590} & 0.702 & 0.805 & \textbf{1.320} & 150 & 172 & \textbf{282} & 0.195 & 0.178 & \textbf{0.120}\\
			\bottomrule
		\end{tabular*}
	\end{center}
	\vspace{-0.15in}
\end{table*}

\begin{table*}[t]
	\small
	\begin{center}
		\caption{Comparison of tracking results with original sequences, random noise, and IoU attack of SiamRPN++~\cite{li-cvpr19-siamrpn++}, DiMP~\cite{bhat-iccv19-learning} and LTMU~\cite{dai-cvpr20-high}  respectively on the VOT2016~\cite{vot2016} dataset.} \label{table:table3}	
		\vspace{0.05in}
		 \begin{tabular*} {17cm} {@{\extracolsep{\fill}}lcccccccccccc}
			\toprule
			\multirow{2}*{\ \ Trackers} & \multicolumn{3}{c}{ Accuracy $\uparrow$ }  & \multicolumn{3}{c}{ Robustness $\downarrow$ } & \multicolumn{3}{c}{ Failures $\downarrow$ } & \multicolumn{3}{c}{ EAO $\uparrow$ } \\
			\cmidrule(r){2-4} \cmidrule(r){5-7} \cmidrule(r){8-10} \cmidrule(r){11-13} 
			~ &Orig. &Rand. &Attack &Orig. &Rand. &Attack &Orig. &Rand. &Attack &Orig. &Rand. &Attack\\
			\midrule
			\ \ SiamRPN++  &0.643 & 0.632 & \textbf{0.605} & 0.200 & 0.340 & \textbf{0.802} &43\ \  &73 \ \ &\textbf{172} &0.461&0.331 &\textbf{0.183}\\
			\ \ DiMP &0.599&0.592 &\textbf{0.536} & 0.140 & 0.168 & \textbf{0.374} &30 \ \ &36\ \ &\textbf{80} \ \  &0.449 & 0.404 &\textbf{0.256}\\
			\ \ LTMU &0.661 &0.646&\textbf{0.604} &0.522&0.592 &\textbf{0.904}&112 &127&\textbf{194} &0.236 & 0.233 & \textbf{0.170}\\
			\bottomrule
		\end{tabular*}
	\end{center}
	\vspace{-0.2in}
\end{table*}

\subsection{Discussions and Visualizations}
\vspace{-0.05in}

In this section, we visualize the variations of adversarial perturbations during IoU attack in Figure~\ref{fig:method}. Given an original image, we iteratively inject the adversarial perturbation as shown in the first row of Figure~\ref{fig:method}. With the increase of adversarial perturbations, the adversarial example drifts the target from the original result and leads IoU scores to decrease. We compute the cosine distance between the perturbations from two consecutive intermediate images. The cosine distance indicates that the generated perturbations follow an increasing trend without fluctuation, decreasing the query numbers effectively in our black-box attack. During each iteration, we visualize the concrete orthogonal composition between the consecutive intermediate images for instance, as shown in the second row of Figure~\ref{fig:method}. We introduce several candidate images according to Eq.~\ref{eq:tan} and select the one with the minimal IoU score as the tangential direction (i.e., \#$1$). Then, we move toward the heavy noise image in trails to make sure the IoU score decreases. We adjust the weight $\epsilon$ in Eq.~\ref{eq:normal} to constrain the variation of perturbation and output the result as the normal direction (i.e., \#$2$). These two directions compose the orthogonal composition during each iteration.  As a result, we hope the final perturbation preserves a lighter degree of noise than heavy random noise does, but the final perturbation can decrease the IoU scores heavily. In other words, our IoU attack makes larger degradation of IoU scores by injecting fewer perturbations.

\section{Experiments}

\renewcommand{\tabcolsep}{0pt}

We validate the performance of our IoU attack on six challenging datasets, 
VOT2019~\cite{vot2019},  VOT2018~\cite{vot2018},  VOT2016~\cite{vot2016}, OTB100~\cite{OTB-2015}, NFS~\cite{nfs} and VOT2018-LT~\cite{vot2018}. Detailed results are provided as follows.

\subsection{Experiment Setup}

{\flushleft\bf Deployment of Trackers.} In order to validate the generality of our black-box adversarial attack, we choose three representative trackers with different structures, SiamRPN++~\cite{li-cvpr19-siamrpn++}, DiMP~\cite{bhat-iccv19-learning} and LTMU~\cite{dai-cvpr20-high}, respectively. SiamRPN++ is a typical detection based tracker with the siamese network. It compares the similarity between a target template and a search region with the region proposal network. The end-to-end learned tracker DiMP exploits both target and background appearance information to locate the target precisely. LTMU is a long-term tracker, utilizing the meta-updater to update the tracker online for target prediction.

{\flushleft\bf Implementation Details.} We formulate the heavy noise image by injecting uniform noise into the clean image as feedback. The type of initial random noise at the same noise level is not sensitive to the degradation of tracking. We discontinue the iterative perturbation update when the IoU score is below the predefined score or the perturbations exceed the maximum. To sum up, the average query numbers of IoU attack are 21.2, 31.4 and 54.2 per frame for SiamRPN++, DiMP and LTMU, respectively. 

\begin{table}
	\small
	\begin{center}
		\caption{Comparison of tracking results with original sequences, random noise, and IoU attack of SiamRPN++~\cite{li-cvpr19-siamrpn++}, DiMP~\cite{bhat-iccv19-learning} and LTMU~\cite{dai-cvpr20-high} respectively on the OTB100~\cite{OTB-2015} dataset.} \label{table:otb}	
		\vspace{0.05in}
		\begin{tabular*} {8.3cm} {@{\extracolsep{\fill}}lcccccc}
		\toprule
		\multirow{2}*{\ \ Trackers} & \multicolumn{3}{c}{ Success $\uparrow$ }  & \multicolumn{3}{c}{ Precision $\uparrow$ } \\
		\cmidrule(r){2-4} \cmidrule(r){5-7}
		~ &Orig. &Rand. &Attack &Orig. &Rand. &Attack\\
		\midrule
		\ \ SiamRPN++ &0.695 & 0.631 & \textbf{0.499} &0.905& 0.818 & \textbf{0.644}\\
		\ \ DiMP &0.671& 0.659 & \textbf{0.592} & 0.869 & 0.860 & \textbf{0.791}\\
		\ \ LTMU  &0.672 & 0.622 & \textbf{0.517} & 0.872 & 0.815 & \textbf{0.712}\\
		\bottomrule
		\end{tabular*}
	\end{center}
	\vspace{-0.2in}
\end{table}

\subsection{Overall Attack Results}

{\flushleft\bf VOT2019.} We implement the three trackers on the VOT2019~\cite{vot2019} dataset consisting of 60 challenging sequences. Different from other datasets, the VOT dataset has a reinitialization module. When the tracker loses the target (i.e., the overlap is zero between the predicted result and the annotation), the tracker will be reinitialized with the ground truth. Failures show the number of re-initialization. Accuracy evaluates the average overlap ratios of successfully tracking frames. Robustness measures the overall lost numbers. In addition, Expected Average Overlap (EAO) is evaluated by a combination of Accuracy and Robustness.

Table~\ref{table:table1} shows the performance drops after IoU attack. We first test all trackers on original sequences. Then we implement our IoU attack method to generate the adversarial examples and evaluate the tracking results. SiamRPN++ leads to more failures than its original results, and the EAO score drops from 0.287 to 0.124. DiMP obtains a 16.5\% drop on its accuracy score, which indicates our attack method leads to an obvious drift. The EAO score also drops dramatically from 0.332 to 0.195. Similarly, our IoU attack method reduces the EAO score of LTMU from 0.201 to 0.150. For further comparison, we also conduct experiments that inject the same level of random noise into the original sequences. Our generated perturbations decrease the IoU scores more dramatically than random noise. 

\begin{table}
	\small
	\begin{center}
		\caption{Comparison of tracking results with original sequences, random noise, and IoU attack of SiamRPN++~\cite{li-cvpr19-siamrpn++}, DiMP~\cite{bhat-iccv19-learning} and LTMU~\cite{dai-cvpr20-high} respectively on the NFS30~\cite{nfs} dataset.} \label{table:nfs}
		\vspace{0.05in}
		\begin{tabular*} {8.3cm} {@{\extracolsep{\fill}}lcccccc}
			\toprule
			\multirow{2}*{\ \ Trackers} & \multicolumn{3}{c}{ Success $\uparrow$ }  & \multicolumn{3}{c}{ Precision $\uparrow$ } \\
			\cmidrule(r){2-4} \cmidrule(r){5-7}
			~ &Orig. &Rand. &Attack &Orig. &Rand. &Attack\\
			\midrule
			\ \ SiamRPN++ & 0.509 & 0.466 & \textbf{0.394} & 0.601& 0.550 & \textbf{0.446}\\
			\ \ DiMP & 0.614 & 0.591 & \textbf{0.545} & 0.729 & 0.710 & \textbf{0.658}\\
			\ \ LTMU  & 0.631 & 0.579 & \textbf{0.462} & 0.764 & 0.699 & \textbf{0.559}\\
			\bottomrule
		\end{tabular*}
	\end{center}
	\vspace{-0.2in}
\end{table}

\def\swseven{0.32\linewidth}
\begin{figure*}[t]
	\begin{center}
		\begin{spacing}{1.4}
			\small
			\begin{tabular}{ccc}
				\includegraphics[width=\swseven]{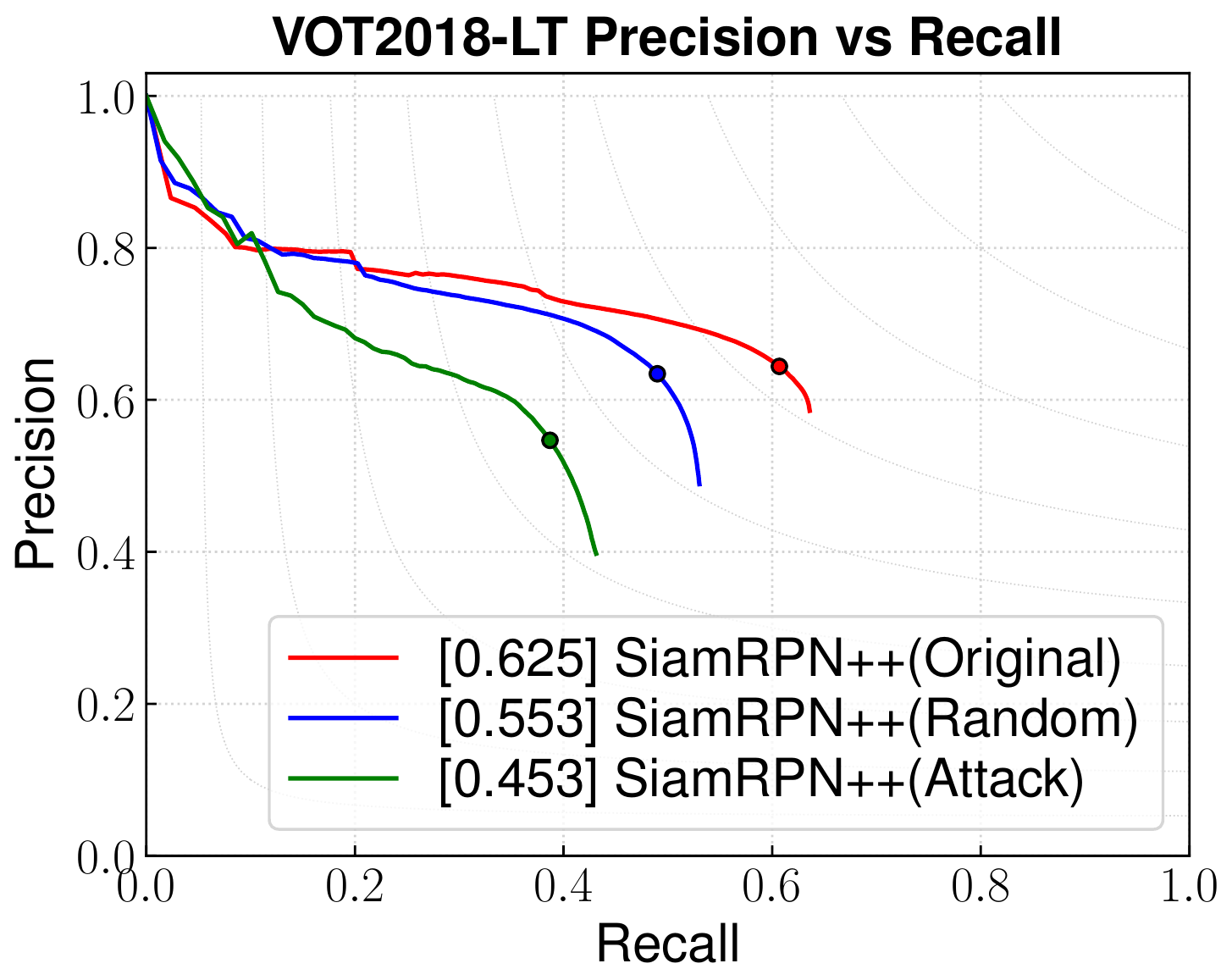} \ \ \ &
				\includegraphics[width=\swseven]{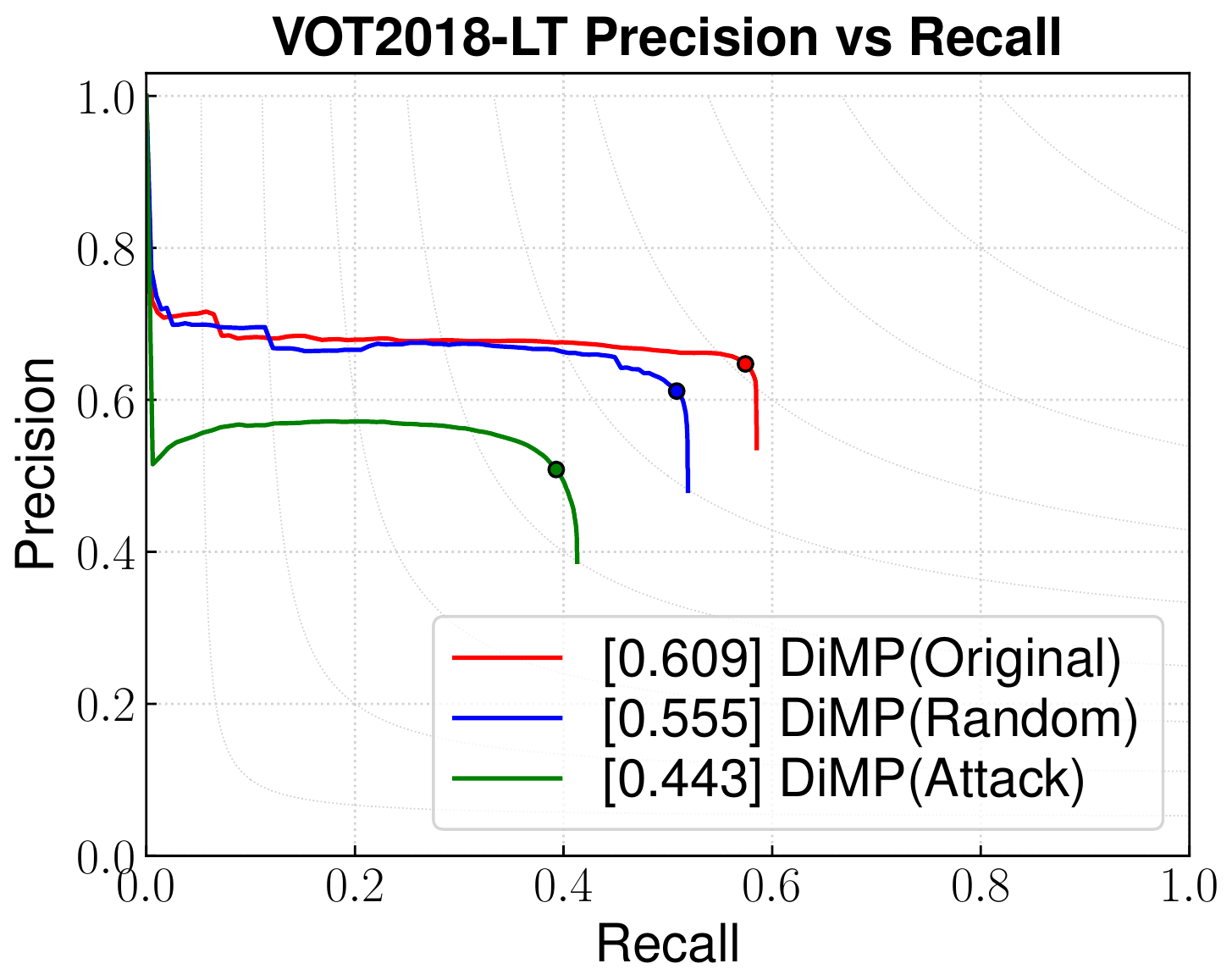} \ \ \ &
				\includegraphics[width=\swseven]{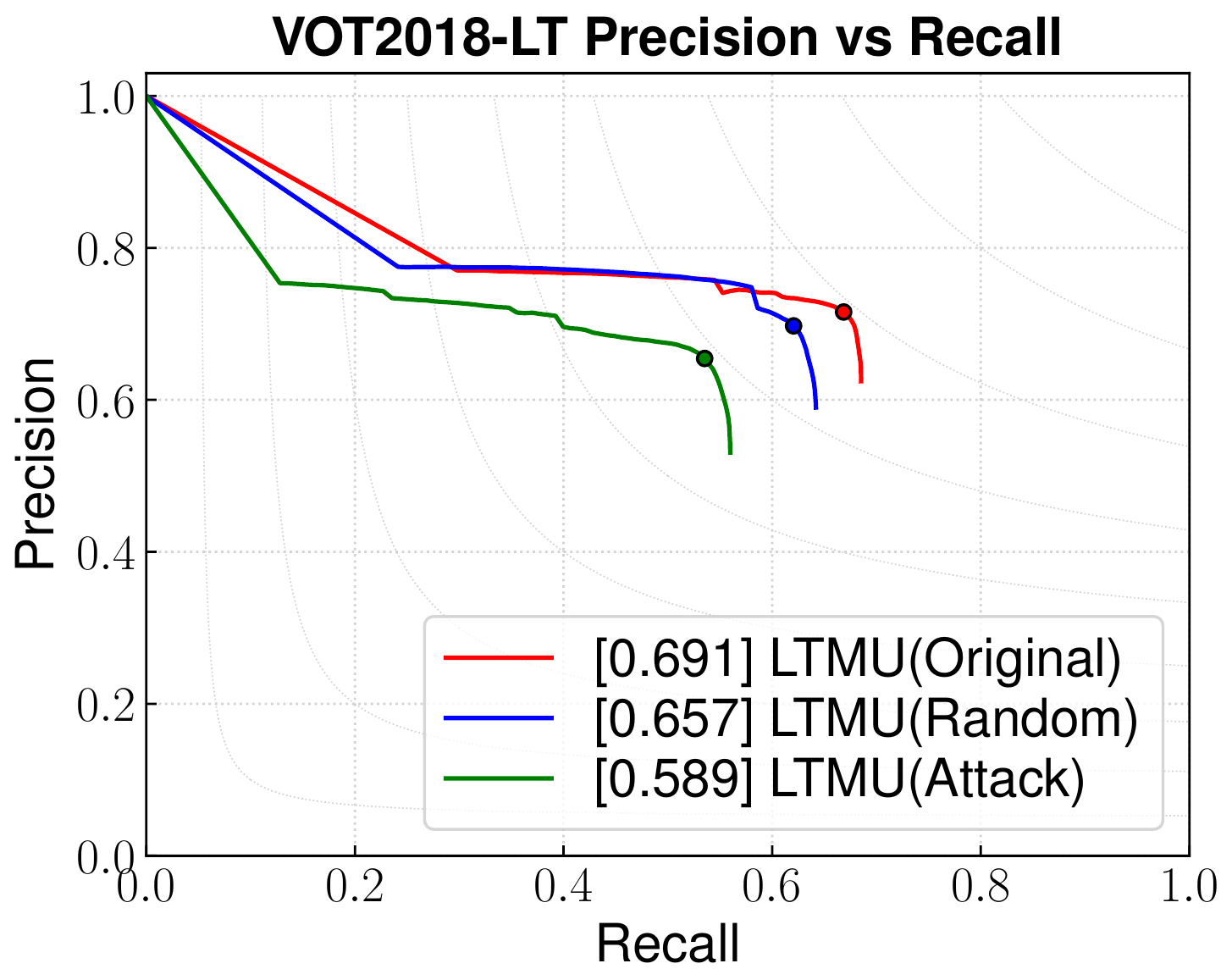}\\
			\end{tabular}
		\end{spacing}
		\vspace{-4mm}
	\end{center}
	\caption{Precision and recall plots of IoU attack for SiamRPN++~\cite{li-cvpr19-siamrpn++},  DiMP~\cite{bhat-iccv19-learning} and LTMU~\cite{dai-cvpr20-high} respectively on the VOT2018-LT dataset~\cite{vot2018}. We use $Attack$ and $Random$ to denote IoU attack and the same level of random noise. The legend is ranked by F-score.} 
	\label{fig:lt}
	\vspace{-3mm}
\end{figure*}

{\flushleft\bf VOT2018.} There are 60 different sequences in the VOT2018~\cite{vot2018} dataset. All the trackers perform favorably on the original sequences. LTMU performs worse than the other two trackers since the re-detection module  yields more reinitializations in VOT-toolkit. Table~\ref{table:table2} shows that the performance of these trackers deteriorates obviously under IoU attack. Concretely, the accuracies of these three trackers get worse after the adversarial attack. These indicate that the trackers indeed deviate from their original results. The primary metric EAO scores are reduced by 68.8\%, 41.9\%, 38.5\% for SiamRPN++, DiMP and LTMU, respectively. 

{\flushleft\bf VOT2016.} Similarly, we also conduct the IoU attack method on the VOT2016 dataset~\cite{vot2016}, as shown in Table~\ref{table:table3}. These trackers perform much better than the above two datasets on the original sequences.  However, IoU attack also reduces the EAO by 60.3\%, 43.0\%, 28.0\% for SiamRPN++, DiMP and LTMU, respectively. Our IoU attack is more effective than the same level of random noise. 

{\flushleft\bf OTB100.} The OTB100~\cite{OTB-2015} dataset includes 100 fully annotated video sequences. The evaluation has two main metrics, success and precision, by using the one-pass evaluation (OPE). We compare the results before and after IoU attack in Table~\ref{table:otb}. With IoU attack, the AUC scores of success significantly decline, accounting for 71.8\%, 88.2\% and 76.9\% of original results for SiamRPN++, DiMP and LTMU, respectively. However, the AUC scores with random noise account for 90.8\%, 98.2\% and 92.6\%, respectively.

{\flushleft\bf NFS30.} We also conduct IoU attack on the NFS30~\cite{nfs} dataset consisting of 100 videos at 30 FPS with an average length of 479 frames. All sequences are manually labeled with nine attributes, like occlusion, fast motion, etc. And we adopt the same metrics used in the OTB100 dataset, as shown in Table~\ref{table:nfs}. According to the AUC metric of success, SiamRPN++ obtains a 22.6\% decrease after IoU attack while injecting the same level of random noise causes an 8.4\% decrease. DiMP achieves an 11.2\% decrease compared to a 3.7\% decrease with random noise. LTMU gets a 26.8\% decrease after IoU attack and an 8.2\% decrease with random noise. IoU attack makes approximately triple drops compared to the same level of random noise.

\def\swsix{0.18\linewidth}
\begin{table}[t]
	\small
	\begin{center}
		\caption{Ablation studies on IoU attack for SiamRPN++~\cite{li-cvpr19-siamrpn++},  DiMP~\cite{bhat-iccv19-learning} and LTMU~\cite{dai-cvpr20-high} on the VOT2018~\cite{vot2018} and VOT2016~\cite{vot2016} datasets. $S_{\rm temporal}$ represents the temporal IoU score and $P_{t-1}$ represents the learned perturbation from historical frames.  } \label{table:ab}
		\vspace{0.05in}
		\begin{tabular*}{8.3cm} {@{\extracolsep{\fill}}lccccc}
			\toprule
			\multirow{2}*{\ \ Tracker}& \multirow{2}*{ $S_{\rm temporal}$}&  \multirow{2}*{ \ \ \ \ \  $P_{t-1}$ \ \ \ \ \ } & \multicolumn{2}{c}{ EAO $\uparrow$}\\
			 \cmidrule(r){4-5} 
			~& ~ &  ~ & \ \ VOT2018 \ \ \ & \ \ VOT2016 \ \ &\\
			\midrule
			\multirow{3}*{\ \ SiamRPN++} & No \  & Yes & 0.149 & 0.189 \\
			~&Yes & No \  & 0.134 & 0.190 \\
			~&Yes & Yes  &\textbf{0.129}& \textbf{0.183} \\
			\midrule
			\multirow{3}*{\ \ DiMP} & No \  & Yes & 0.257 & 0.275 \\
			~& Yes & No \  & 0.261 & 0.295 \\
			~& Yes & Yes &\textbf{0.248}& \textbf{0.256}\\
			\midrule
			\multirow{3}*{\ \  LTMU}  &No \  & Yes &0.147& 0.184 \\
			~& Yes & No \  & 0.150 & 0.189 \\
			~& Yes & Yes &\textbf{0.120} & \textbf{0.170} \\
			\bottomrule
		\end{tabular*}
	\end{center}
	\vspace{-0.2in}
\end{table}

{\flushleft\bf VOT2018-LT.} In order to further verify the effectiveness of our IoU attack, we conduct three trackers on a more challenging dataset VOT2018-LT~\cite{vot2018}. It has 35 sequences with an average length of 4200 frames, which is much longer than other datasets and closer to practical applications. Each tracker needs to output a confidence score for the target being present and a predicted bounding box in each frame. Precision ($P$) and recall ($R$) are evaluated for a series of confidence thresholds, and the F-score is calculated as $F$ = $ 2 P \cdot R/ (P + R)$. The primary long-term tracking metric is the highest F-score among all thresholds. Figure~\ref{fig:lt} shows the results of precision and recall at different confidence thresholds before and after IoU attack. The results in the legend are ranked by F-score. The precision and recall both drop significantly after our IoU attack on three trackers. Our IoU attack method reduces the F-Score by 27.5\%, 27.3\% and 14.8\% for SiamRPN++, DiMP and LTMU, respectively. All trackers after IoU attack perform poorly compared with injecting the same level of random noise. Our black-box attack method is proven to be also effective for long-term tracking. 

\subsection{Ablation Studies}

\begin{figure}[t]
	\centering
	\includegraphics[width=0.90\linewidth]{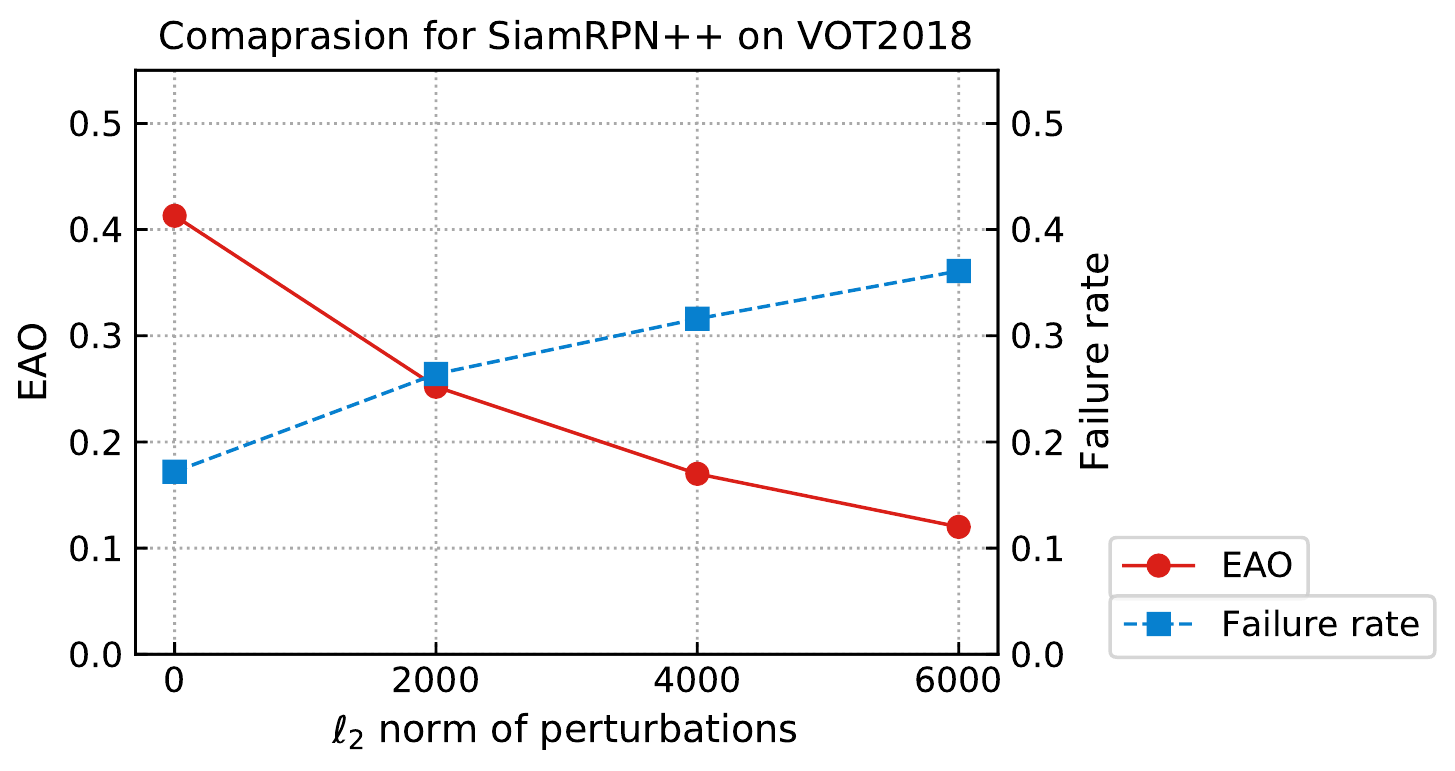}
	\vspace{-1mm}
	\caption{Comparison on EAO scores and failure rates of different perturbations for SiamRPN++~\cite{li-cvpr19-siamrpn++} on the VOT2018~\cite{vot2018} dataset.
	} \label{fig:ab}
	\vspace{-0.05in}
\end{figure}

To explore the temporal motion of visual object tracking in black-box attack, we separately compare the IoU attack method with or without involving temporal IoU scores in Eq.~\ref{eq:score}, as reported in Table~\ref{table:ab}. With the help of temporal IoU scores, the deep trackers get worse tracking accuracies than only using the spatial IoU scores on multiple datasets. In addition, we transfer the learned perturbation $P_{t-1}$ into the next few frames and use its weight to initialize the input for temporally consistent motion attack. We compare the IoU attack method with or without the transfer of historical perturbation $P_{t-1}$ in Table~\ref{table:ab}. The overall performance metric EAO indicates that the transfer and initialization of previous perturbation indeed improve the attack effects and decrease the tracking accuracies. In addition, we also illustrate the attack performance with the variation of perturbations on the VOT2018~\cite{vot2018} dataset, as shown in Figure~\ref{fig:ab}. The perturbations are measured by $\ell_2$ norm. Failure rate represents the average rate of failure frames in the whole video. We observe that the attack performance gets worse with the increase of perturbations accordingly.  

\begin{table}[t]
	\small
	\begin{center}
		\caption{ Comparison with existing white-box and black-box attack methods for SiamRPN++~\cite{li-cvpr19-siamrpn++} with ResNet on the OTB100~\cite{OTB-2015} dataset. } \label{table:comparison}
		\vspace{0.05in}
		\begin{tabular*}{8.3cm} {@{\extracolsep{\fill}}lccc}
			\toprule
			\ Method & Succ. Drop & Prec. Drop&Type\\
			\midrule
			\ CSA~\cite{yan-cvpr20-cooling}& 37.2 &  44.3& White-box\\
			\ SPARK~\cite{guo-eccv20-spark} & \ \ 6.6 &\ \ 2.7 & Black-box\\
			\ UAP~\cite{moosavi-iccv17-uap} &\ \ 2.7  &\ \ 4.0 & Black-box\\
			\ Ours  &\textbf{19.6} &\textbf{26.1}& Black-box \\
			\bottomrule
		\end{tabular*}
	\end{center}
	\vspace{-0.2in}
\end{table}

\subsection{Comparison with Other Methods}

Table~\ref{table:comparison} reports the comparison with existing white-box and black-box attack methods. Our black-box attack method without access to the network architecture of trackers performs slightly worse than the white-box attack method CSA~\cite{yan-cvpr20-cooling} for tracking. SPARK~\cite{guo-eccv20-spark} performs the transfer-based black-box attack and obtains a 6.6\% success drop and a 2.7\% precision drop. Our decision-based black-box attack significantly outperforms SPARK. In addition, we apply the perturbation from UAP~\cite{moosavi-iccv17-uap} frame by frame, which is designed for attacking the classification in static images. Our method considering the temporal motion changes of the target objects achieves much greater success. 

\subsection{Qualitative Results}

Figure~\ref{fig:qua} qualitatively shows the tracking results of our IoU attack for SiamRPN++~\cite{li-cvpr19-siamrpn++},  DiMP~\cite{bhat-iccv19-learning} and LTMU~\cite{dai-cvpr20-high} on three challenging sequences. We visualize the original tracking results in (a), the results with the same level of random noise in (b) and the results of our IoU attack in (c). In the original images, all these trackers locate the target objects and estimate the scale changes accurately. After generating the adversarial examples, these trackers estimate the target location inaccurately. However, the same level of random noise cannot drift the trackers, as shown in the second column. This indicates that the proposed IoU attack generates the optimized perturbations and maintains the same level of random noise. 


\def\swthree{0.33\linewidth}
\begin{figure}[t]
	\begin{center}
		\includegraphics[width=0.8\linewidth]{\figdir/intro/legend_3.pdf}\\
	\end{center}
	\vspace{-0.35in}
	\begin{center}
		\small
		\begin{tabular}{ccccc}
			\includegraphics[width=\swthree]{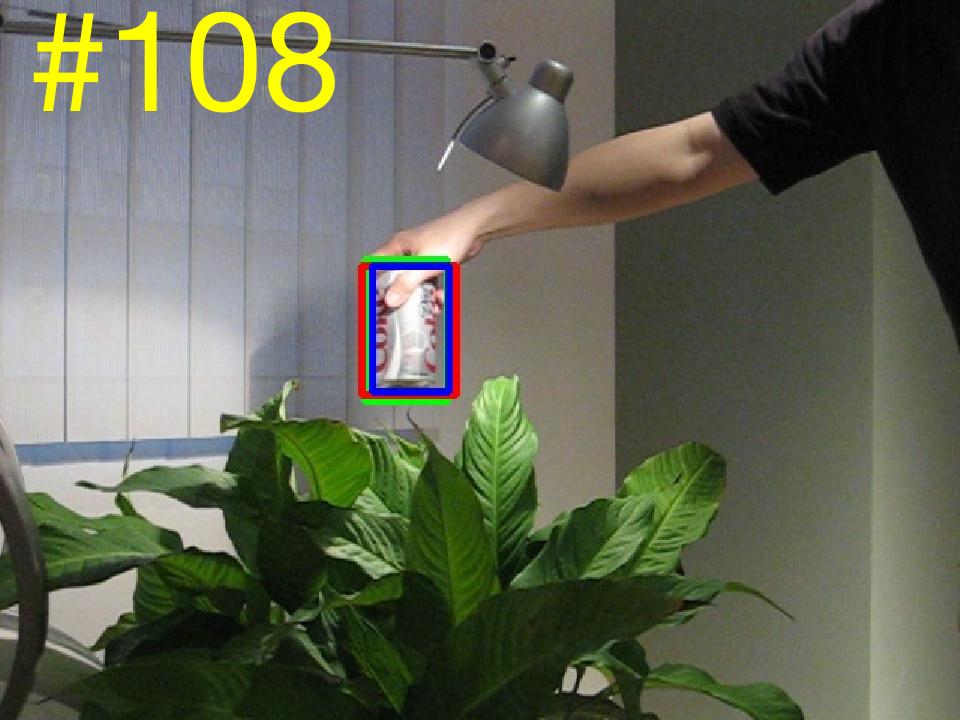}\ \ &
			\includegraphics[width=\swthree]{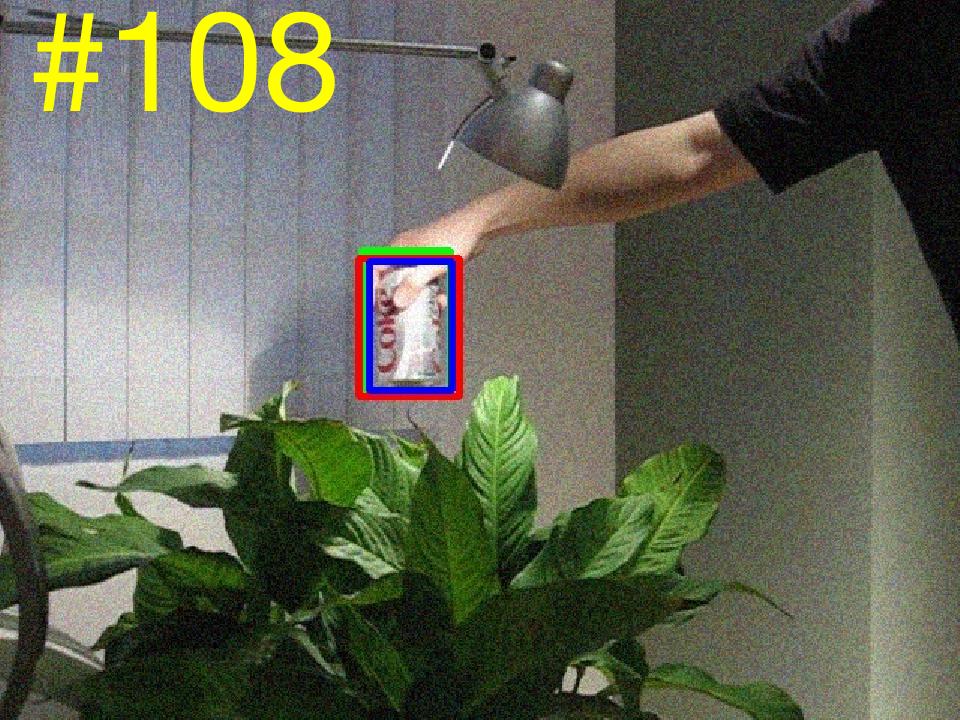}\ \ &
			\includegraphics[width=\swthree]{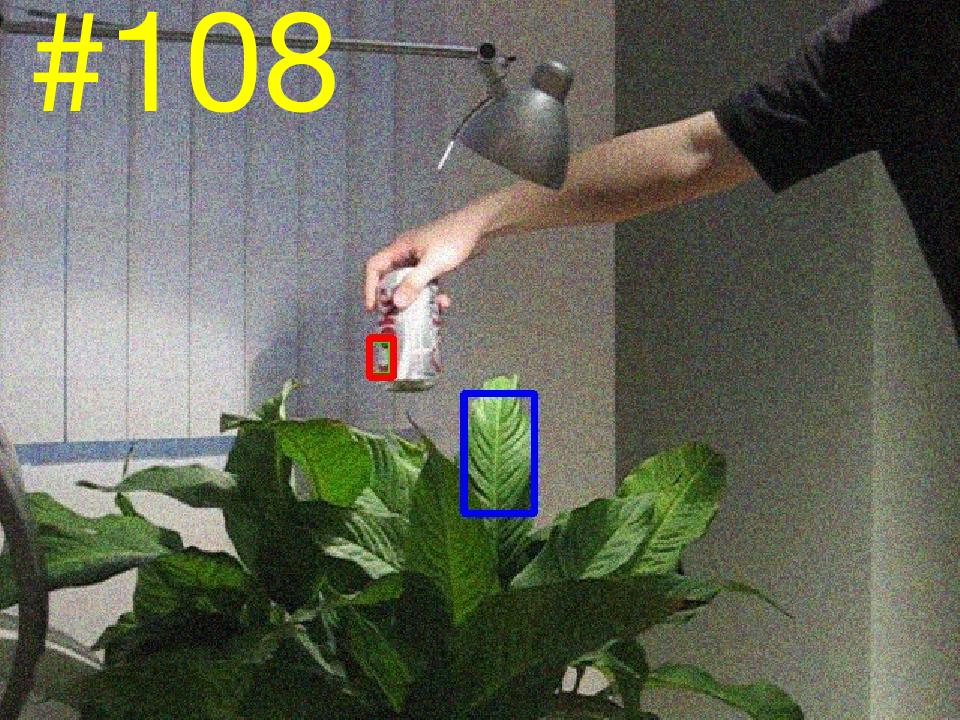}\ &
			\\
			\includegraphics[width=\swthree]{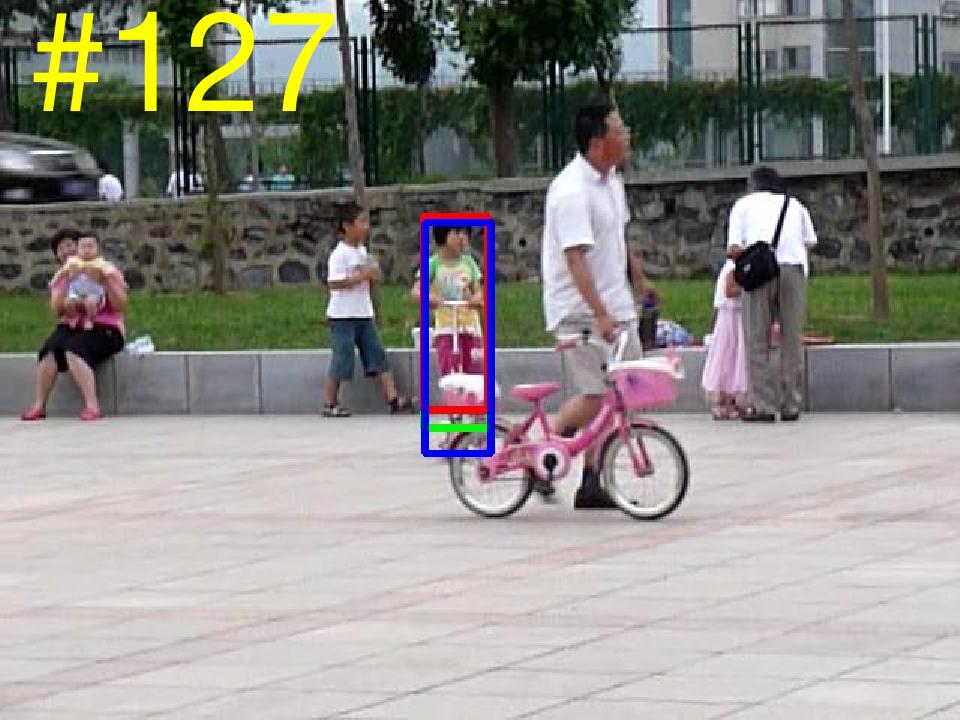}\ \ &
			\includegraphics[width=\swthree]{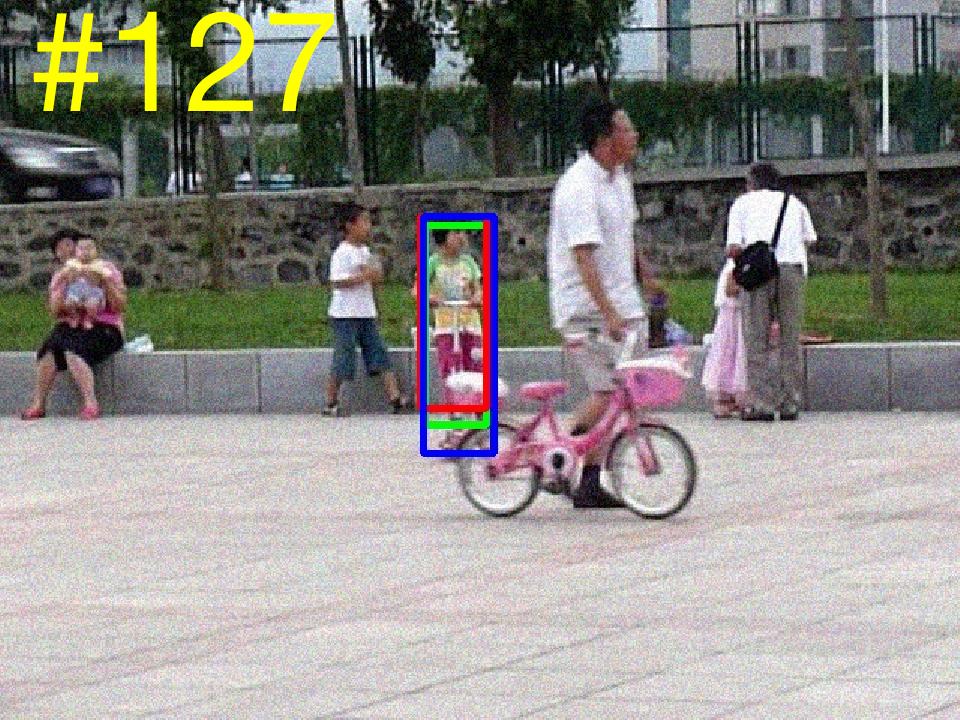}\ \ &
			\includegraphics[width=\swthree]{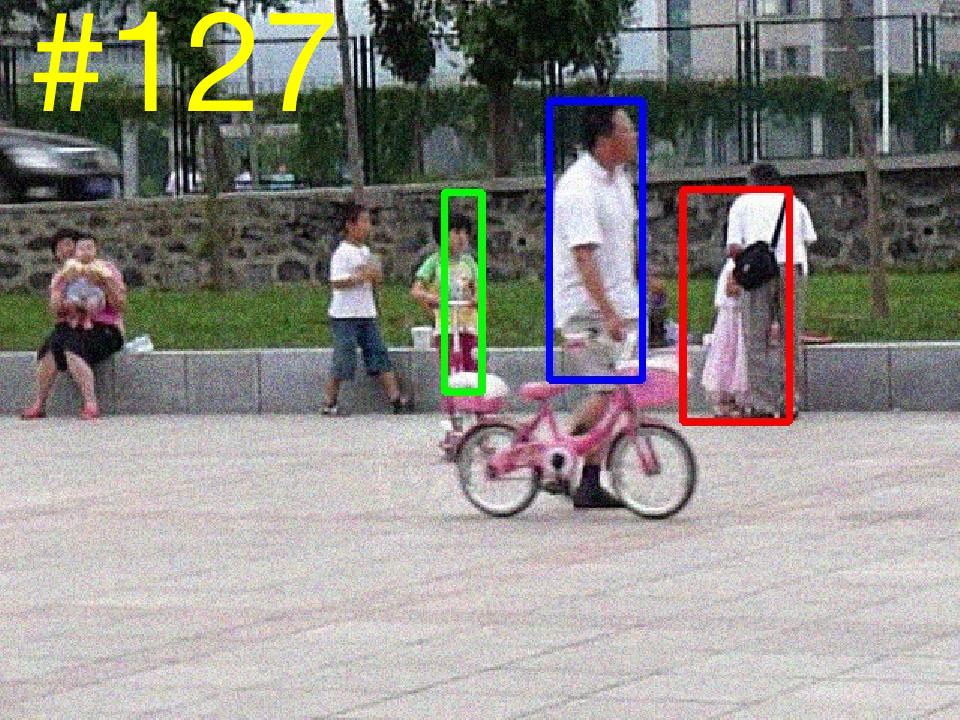}\ &
			\\
			\includegraphics[width=\swthree]{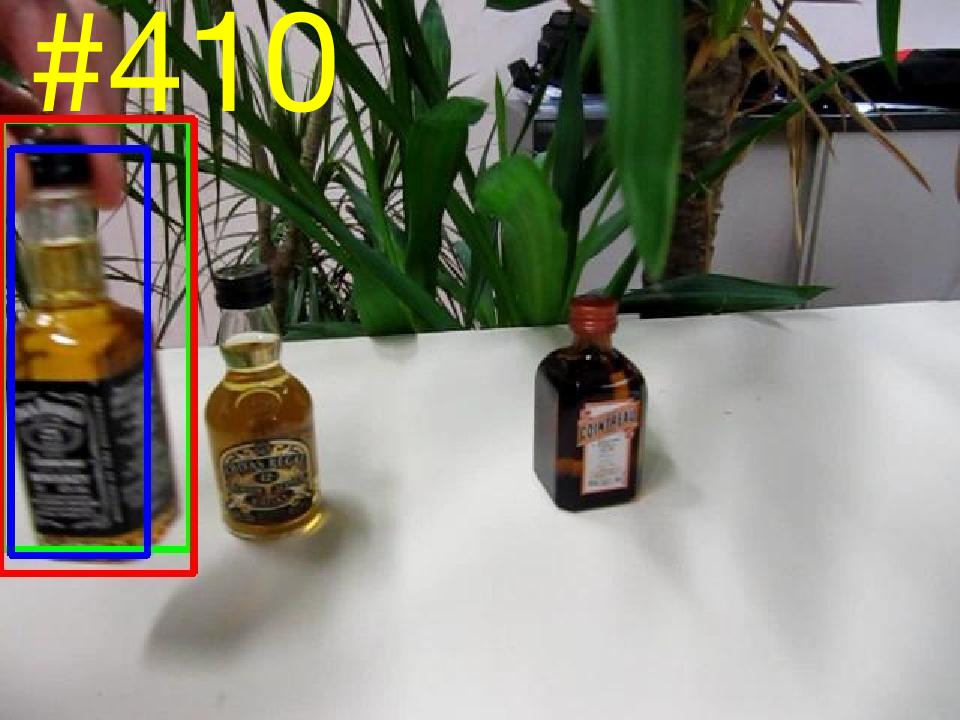}\ \ &
			\includegraphics[width=\swthree]{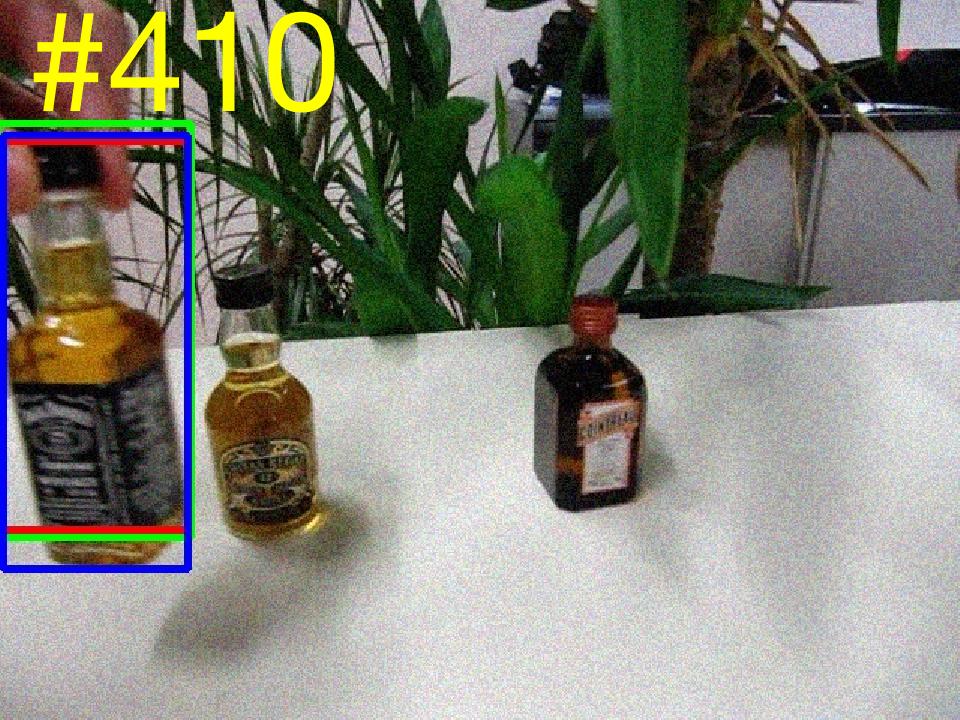}\ \ &
			\includegraphics[width=\swthree]{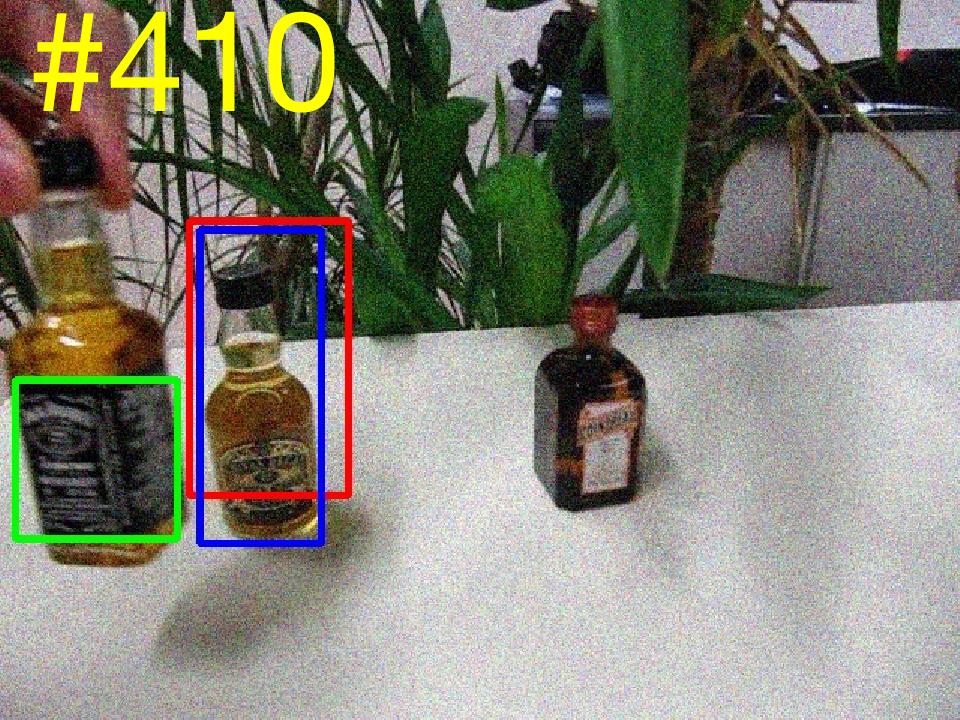}\ &
			\\
		\end{tabular}
		\begin{tabular} {ccc}
			(a) Original results \ \ \ \ \ &  (b) Random results & \ \ \ \ \ \ \ (c) Attack results \\
		\end{tabular}
	\end{center}
	\vspace{-0.1in}
	\caption{Qualitative results of IoU attack on three challenging sequences from the OTB100~\cite{OTB-2015} dataset.} 
	\label{fig:qua}
\end{figure}

\section{Concluding Remarks}

\renewcommand{\tabcolsep}{0pt}

In this paper, we propose an IoU attack method in the black-box setting to generate adversarial examples for visual object tracking. Without access to the network architecture of deep trackers, we iteratively adjust the direction of light-weight noise according to the predicted IoU scores of bounding boxes, which involve temporal motion in historical frames. Furthermore, we transfer the perturbations into the next frames to improve the effectiveness of attack. We apply the proposed method to three state-of-the-art representative trackers to illustrate the generality of our black-box adversarial attack for visual object tracking. The extensive experiments on standard benchmarks demonstrate the effectiveness of the proposed black-box IoU attack. We believe this work helps to evaluate the robustness of visual object tracking.\\

\vspace{-0.05in}

\noindent \textbf{Acknowledgements.} This work was supported by NSFC (61906119, U19B2035), Shanghai Municipal Science and Technology Major Project (2021SHZDZX0102), and Shanghai Pujiang Program. 



\clearpage
{\small
\bibliographystyle{ieee_fullname}
\bibliography{ref}
}

\end{document}